\documentclass[letterpaper, 10 pt, journal, twoside]{IEEEtran}

\usepackage{amsmath}
\usepackage{amsfonts}
\usepackage{amssymb}

\usepackage{amsthm}
\usepackage[ruled,vlined]{algorithm2e}
\usepackage{mathtools}
\usepackage{tabularx}
\usepackage{graphicx}
\usepackage{subfigure}
\usepackage{enumerate}
\usepackage{float}
\usepackage{url}
\usepackage{verbatim}
\usepackage{stfloats}
\usepackage{bbding}
\usepackage{bm}
\usepackage{hyperref}
\usepackage{booktabs}
\usepackage{threeparttable}  
\usepackage{multirow}
\usepackage{pifont}  
\usepackage{subcaption}
\usepackage{colortbl}
\usepackage{array}
\usepackage{booktabs}
\usepackage{cite}
\usepackage[table,dvipsnames]{xcolor} 

\graphicspath{{figures/}}
\IEEEoverridecommandlockouts

\usepackage[normalem]{ulem}

\begin{document}

\title{Drive in Corridors: Enhancing the Safety of\\ End-to-end Autonomous Driving\\ via Corridor Learning and Planning }

\author{Zhiwei Zhang$^{1}$,
        Ruichen Yang$^{1}$,
        Ke Wu$^{1}$,
        Zijun Xu$^{1}$, \\
        Jingchu Liu$^{2}$,
        Lisen Mu$^{2}$,
        Zhongxue Gan$^{1}$
        and Wenchao Ding$^{1}$ 
\thanks{Manuscript received: January, 2, 2025; Revised April, 2, 2025; Accepted April, 24, 2025.}
\thanks{This paper was recommended for publication by Editor Pascal Vasseur upon evaluation of the Associate Editor and Reviewers' comments.
This work was supported by the Science and Technology Commission of Shanghai Municipality (No. 24511103100), National Natural Science Foundation of China (NSFC) under Grant 62403142 and Shanghai Pujiang Program (23PJ1400900). \textit{(Corresponding author: Wenchao Ding).}} 
\thanks{
$^{1}$Zhiwei Zhang, Ruichen Yang, Ke Wu, Zijun Xu, Zhongxue Gan and Wenchao Ding are with College of Intelligent Robotics and Advanced Manufacturing, Fudan University, China.
        {\tt\footnotesize zhiweizhang23@m.fudan.edu.cn, dingwenchao@fudan.edu.cn}}%
\thanks{$^{2} $ Jingchu Liu and Lisen Mu are with Horizon Robotics.
       }%
\thanks{Digital Object Identifier (DOI): see top of this page.}
}

\markboth{IEEE Robotics and Automation Letters. Preprint Version. Accepted April, 2025}
{Zhang \MakeLowercase{\textit{et al.}}: Drive in Corridors}

\maketitle


\begin{abstract}
Safety remains one of the most critical challenges in autonomous driving systems.
In recent years, the end-to-end driving has shown great promise in advancing vehicle autonomy in a scalable manner.
However, existing approaches often face safety risks due to the lack of explicit behavior constraints.
To address this issue, we uncover a new paradigm by introducing the \textit{corridor} as the intermediate representation.
Widely adopted in robotics planning, the corridors represents spatio-temporal obstacle-free zones for the vehicle to traverse.
To ensure accurate corridor prediction in diverse traffic scenarios, we develop a comprehensive learning pipeline including data annotation, architecture refinement and loss formulation.
The predicted corridor is further integrated as the constraint in a trajectory optimization process.
By extending the differentiability of the optimization, we enable the optimized trajectory to be seamlessly trained within the end-to-end learning framework, improving both safety and interpretability.
Experimental results on the nuScenes dataset demonstrate state-of-the-art performance of our approach, showing a 66.7\% reduction in collisions with agents and a 46.5\% reduction with curbs, significantly enhancing the safety of end-to-end driving. Additionally, incorporating the corridor contributes to higher success rates in closed-loop evaluations.
Project page: \url{https://zhiwei-pg.github.io/Drive-in-Corridors}.
\end{abstract}


\section{Introduction}
\label{sec:Introduction}

\IEEEPARstart{E}{nd-to-end} autonomous driving has recently gained attention as a promising alternative to traditional methods, offering better scalability and adaptability to complex real-world scenarios. 
This approach directly maps raw sensor data to trajectories through a unified neural network trained to imitate human drivers. 
By integrating modules (e.g. perception, prediction, and planning) into a single trainable architecture, it eliminates the need for handcrafted rules.

Despite the significant promise of end-to-end methods, ensuring safe actions remains a critical challenge.
This challenge arises from the inherent lack of precise mathematical guarantees and interpretability in learning-based approaches\cite{chen2024end}.
Previous studies\cite{sadat2020perceive, casas2021mp3, chen2024vadv2} seek to address this issue by introducing safety cost functions over sampled trajectories. 
However, this strategy relies on a large quantity of high-quality trajectory samples and precise evaluation functions. 
UniAD\cite{hu2023planning} incorporates post trajectory optimization on occupancy grids, while VAD\cite{jiang2023vad} adopts constraint losses to enhance trajectory safety. 
While these approaches have shown some success, there remains significant room for improving the safety of end-to-end driving systems.

To resolve the above challenges and elevate the safety of end-to-end autonomous driving, we undertake two primary investigations.
The first involves tackling the absence of an effective representation to constrain the vehicle’s actions.
To this end, we introduce the core concept of corridor into our work.
Safe corridor, a widely used tool in robotics planning, is a geometric representation that defines safe, obstacle-free regions the vehicle is going to traverse.
We integrate corridor learning into the multi-task pipeline of end-to-end driving by annotating corridors in the dataset, designing network architecture and devising loss functions. 
These components empower the model to accurately and flexibly identify safe drivable area across diverse and dynamic driving scenarios.

Second, we adopt a trajectory optimization process to generate the planning trajectory.
Compared to neural-based planning heads \cite{hu2023planning, chen2024vadv2, jiang2023vad ,liao2024diffusiondrive}, our approach offers a distinct advantage in interpretability.
Previously overlooked considerations such as vehicle kinematics and control bounds are inherently incorporated into the optimization formulation.
Moreover, as a representation of intended driving zones, the corridor integrates naturally with the optimization by enforcing constraints on the vehicle's state.
Furthermore, leveraging advances in optimization theories \cite{amos2017optnet}, we ensure that the optimization process is differentiable, allowing the gradient of the solution to propagate back through the network.
Consequently, the optimized trajectory can be seamlessly trained within the joint learning framework, marking a broadened scope of end-to-end driving while fostering a more cohesive planning process.

Bringing everything together, we present \textbf{CorDriver}, an end-to-end driving model with explicit and differentiable safety constraints. 
Our design is implemented on top of the leading-edge model VAD \cite{jiang2023vad} and rigorously evaluated with respect to collisions with agents and curbs.
On the nuScenes dataset \cite{caesar2020nuscenes}, our approach achieves a 66.7\% reduction in agent collisions and a 46.5\% reduction in curb collisions.
Similar improvements are observed in the Bench2Drive\cite{jia2024bench2drive} closed-loop tests, where integrating corridor into the planning process leads to higher success rate.
These advancements represent a significant step toward making end-to-end autonomous driving both safer and more interpretable.

We summarize the contributions of this paper as follows:
\begin{itemize}
\item[1)] We propose an explicit and interpretable approach to enhance the safety of autonomous vehicles within the end-to-end framework.
\item[2)] To the best of our knowledge, we are the first to introduce the safe corridor into learning based autonomous driving. We develop a complete pipeline for corridor learning and demonstrate its effectiveness in improving driving safety.
\item[3)] A differentiable optimization process incorporating the corridor as the constraint enables the generation of safe trajectories while considering vehicle kinematics, thereby enhancing the interpretability of end-to-end driving.
\item[4)] Through sufficient and comprehensive validations, our approach demonstrates a significant improvement in the safety of end-to-end planning.
\end{itemize}

\begin{figure*}[htbp]
	\centering
	\includegraphics[width=\textwidth]{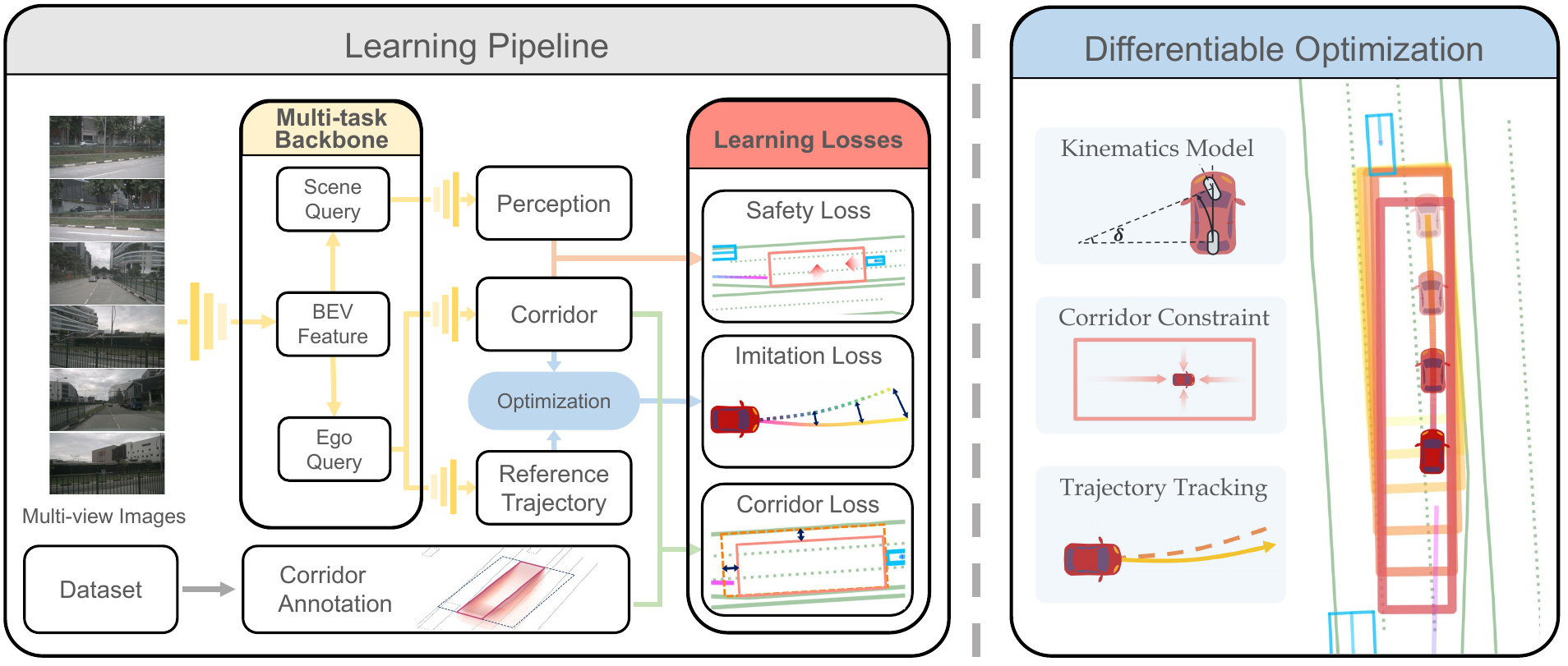}
	\caption{Architecture of our method. The multi-task backbone processes multi-view images as input and outputs perception results, the reference trajectory, and the corridor. The predicted corridor is supervised using the annotation from the dataset and further refined through the safety loss to minimize overlap with agents and curbs. Built on the kinematic bicycle model, the differentiable optimization module utilizes the corridor as the constraint and is aimed to track the reference trajectory. Finally, the optimized trajectory is trained to imitate human driving actions.}
        \vspace{-0.3cm}
	\label{fig:system_architecture}
\end{figure*}


\section{Related Work}
\label{sec:RelatedWork}

\subsection{Safe End-to-end Driving}

A common approach to ensure safety in end-to-end autonomous driving is to assign cost functions to a set of sampled trajectories\cite{sadat2020perceive}\cite{casas2021mp3}\cite{chen2024vadv2}.
Earlier methods \cite{sadat2020perceive} \cite{casas2021mp3} predict occupancy grids and count the number of overlapping cells with the ego vehicle as a safety evaluation metric. 
VADv2 \cite{chen2024vadv2} focuses on modeling trajectory distributions from demonstrations, penalizing colliding samples with lower probabilities. 
Although effective, these sampling-based methods are highly dependent on the quality and diversity of the sampled trajectories, and a larger trajectory library inevitably increases computational overhead.

Other end-to-end approaches predict trajectories directly through regression, such as UniAD \cite{hu2023planning} and VAD \cite{jiang2023vad}. 
UniAD refines trajectories through post-optimization to repel them from occupancy grids, while VAD introduces additional loss functions based on vectorized perception outputs to penalize collisions. 
More recent works \cite{zhai2023rethinking}\cite{Li_2024_CVPR} highlight the connection between trajectory quality and ego status, while others \cite{liao2024diffusiondrive}\cite{sun2024sparsedrive}\cite{weng2024drive} propose improved architectures and decoders, advancing trajectory planning to unprecedented levels of precision. 
Despite these advancements, ensuring safety remains a major challenge in end-to-end trajectory planning.

\subsection{Safe Corridor}

Safe corridor is a powerful representation for collision avoidance and trajectory optimization in robotics planning. 
It is first introduced in robotics planning through IRIS \cite{deits2015computing}, where convex obstacle-free regions are computed through iterative optimizations to plan the footsteps of bipedal robots. 
Liu et al. \cite{liu2017planning} enhance the efficiency of this approach and successfully apply the corridor to a trajectory optimization problem by formulating the constraint as linear inequalities, limiting trajectories inside polyhedrons. 
The concept of safe corridor has been adapted to autonomous driving in the subsequent works\cite{li2021optimization}\cite{han2023efficient}\cite{ding2019safe}.
In this work, we seek to extend the use of this powerful representation to the domain of end-to-end driving.
The architecture is demonstrated in Fig. \ref{fig:system_architecture}.  


\section{Corridor Learning}
\label{sec:corridor}


\subsection{Corridor Representation}
Corridors are commonly represented using connected polyhedrons or spheres. 
However, given the highly structured nature of driving environments, such as lanes and agent bounding boxes, a simpler yet effective choice is to represent the corridor using rectangles. 
These rectangles can be adequately described on the 2D Bird's Eye View (BEV) using properties such as position, orientation, and size.
To incorporate temporal information in driving scenes, we assign one rectangle to each planning timestamp, resulting in the corridor formulation $\mathcal{C}$:
\begin{equation}
    \mathcal{C} = \{ \mathbf{c}_t = \left[c_x,c_y,\theta, l,w \right]_t \Big| t=1,...N \},
\end{equation}
where $c_x$ and $c_y$ denote the rectangle's center position, $\theta$ represents its heading, and $l$ and $w$ correspond to its length and width. The variable $N$ indicates the total number of future timestamps.

In practice, we find this representation effective at capturing safe regions, even in highly dynamic scenarios
and on curved roads. Compared to alternatives like 4D occupancy, it is more lightweight and better suited for the subsequent convex optimization.
In the rest of the paper, the term \textit{rectangle} refers to the extracted empty space at one timestamp, and the union of rectangles over the planning horizon makes up the \textit{corridor}.

\subsection{Corridor Annotation}
\label{sec: ann}

\begin{figure*}[htb]
	\centering
	\includegraphics[width=1.0\linewidth]{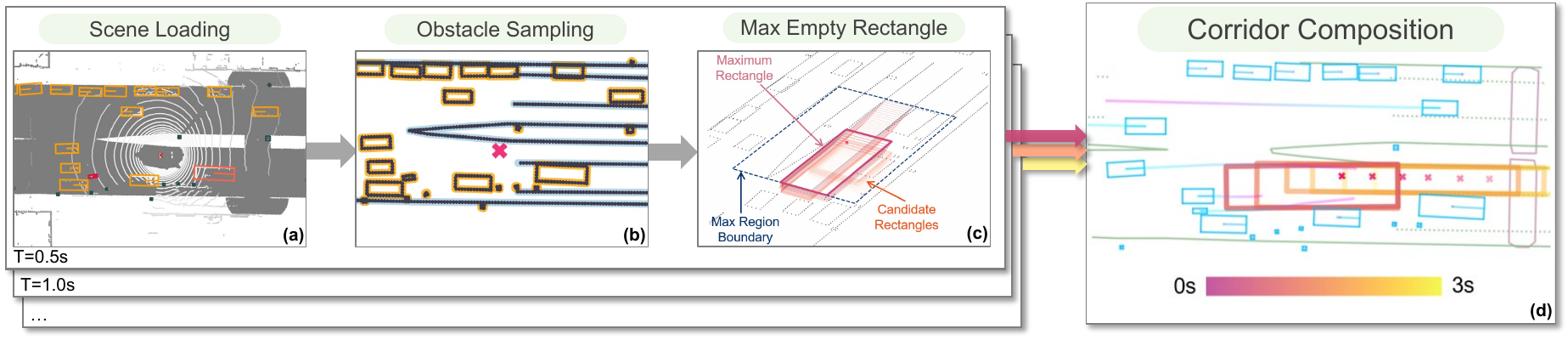}
	\captionsetup{font={small}}
	\caption{Corridor Annotation. (a) Agent boxes and map data are loaded from the dataset. (b) Coordinates are sampled. (c) The largest rectangle is identified and highlighted on top and other candidate rectangles are displayed below ground level with high opacity. (d) Iterating over timestamps and combining the results forms the corridor, where rectangles are stacked on the BEV, and colors represent different timestamps.}
        \vspace{-0.5cm}
	\label{fig:corridor}
\end{figure*}

To enable the model to identify safe corridors, we need to generate corridor annotations as supervisions. 
The first step to label corridors in the dataset is to identify both physical and semantic obstacles in the traffic scene. 
We take the following concerns for obstacle selection.
\textbf{Agents}. Traffic participants such as vehicles and pedestrians are critical for safe driving, and their bounding boxes are maintained as obstacles. 
\textbf{Curbs}. Road boundaries act as essential constraints, ensuring the vehicle remains within drivable areas. 
\textbf{Lanes}. Lanes also serve as implicit restrictions to guide driving behavior. To this end, we first extract the ego trajectories in a recent period $T_{ego}$, and keep lanes that do not overlap with the trajectory as flexible semantic obstacles.
Based on the geometry of these elements, we sample 2D coordinates along their contours using a specified threshold $\delta_{obs}$, forming the obstacle point set: 
\begin{equation}
 \mathcal{O}=\left\{ [p^o_x, p^o_y]_i, \Big| i=1,...N_o  \right\},
\end{equation}
where $N_o$ is the total number of points.
Notably, this approach is adaptable and can be extended to additional semantics, such as traffic lights and pedestrian crossings.

With the essential obstacles identified, we can now generate the safe corridor by finding the maximum rectangles.
At each timestamp $t$, the 2D obstacle points $\mathcal{O}_t$ are extracted, and their coordinates are transformed into the local frame based on the ego state $\left(p_x^{ego}, p_y^{ego}, \theta^{ego} \right)$.
To enhance computational efficiency, we define a maximum region with dimensions $\left(l_{max}, w_{max}\right)$ and disregard points outside the region.
We fix the rectangle's orientation to match the ego's heading $\theta=\theta^{ego}$, and focus on determining the optimal center position $\left(c_x, c_y\right)$ and size $\left(l, w\right)$ of the rectangle.
The problem is formulated as the well-studied Maximum Empty Rectangle (MER) problem\cite{naamad1984maximum} in computational geometry —that is, finding the largest axis-aligned rectangle within a given boundary that contains an origin while avoiding obstacle points. To solve the MER problem, we traverse combinations of obstacle points, where each point precisely defines a candidate rectangle's edge. The area of each valid rectangle is computed and compared, and the largest rectangle is selected.
Repeating this process at each timestamp and composing the rectangles together produces the complete annotation for the corridor. 
An illustration of the process is shown in Fig. \ref{fig:corridor}.

\subsection{Corridor Learning}
Our end-to-end learning model is based on VAD \cite{jiang2023vad}, which utilizes vectorized scene representations and employs transformer-based interactions to produce detections, predictions, and maps from the image inputs.
Its planning head takes the interacted ego query $Q_{ego}$, along with the ego status and driving commands, and decodes the ego trajectory $\hat{\boldsymbol{\xi}}$ using an MLP.
In the following sections, we refer to this trajectory as the \textit{reference trajectory}.
Inspired by the observation that corridors are constructed around an initial path indicating ego intent\cite{deits2015computing}\cite{liu2017planning}, we introduce an additional head parallel to the original planning head, directly decoding corridor predictions from $Q_{ego}$.

We supervise the predicted corridor using the annotations from Section \ref{sec: ann}.
Following prior works \cite{carion2020end}\cite{liao2024maptrv2}, the corridor orientations are represented using their cosine and sine components. 
The corridor loss is computed as 
\begin{equation}
\label{eq: cor}
\begin{aligned}
    \mathcal{L}_{cor} & = \mathcal{L}_1\left(\hat{\mathcal{C}}, \mathcal{C}^{\star} \right),
\end{aligned}
\end{equation}
where $\mathcal{L}_1$ is the L1 loss, $\hat{\mathcal{C}}$ and $\mathcal{C}^{\star}$ are predicted and ground truth corridors respectively.

To further capture the geometrical properties, we introduce three auxiliary losses for corridor learning, including two safety loss terms that repel the corridor from obstacles. 
The perception outputs provide the locations of curbs and agents. 
Curbs are represented as a set of points, $\mathcal{M}$, marking the road boundaries, while agents are simplified to the vertices of their bounding boxes, $\mathcal{A}$, obtained from the detection and prediction head.
Note that the safety losses do not penalize overlaps with lane dividers. As a result, the corridor predictions may intersect with some lane markers in favor of a larger safe area, imitating the human tendency to utilize adjacent lane space when driving.
Subsequently, the edges of each rectangle of the predicted corridor are computed, and the safety losses are determined by the minimum distance between an obstacle point and the edges.

Specifically, the map safety loss $\mathcal{L}_{map}$ and the agent safety loss $\mathcal{L}_{agent}$ are formulated as,
\begin{equation}
\label{eq: map}
\begin{aligned}
    \mathcal{L}_{map} = \sum_{t=1}^{N} \max_{i} \mathbb{D}\left(\mathbf{p}^i_t, \mathbf{c}_t \right), \mathbf{p}^i_t \in \mathcal{M} ,
\end{aligned}
\end{equation}

\begin{equation}
\label{eq: agent}
\begin{aligned}
    \mathcal{L}_{agent} = \sum_{t=1}^{N} \max_{i} \mathbb{D}\left(\mathbf{p}^i_t, \mathbf{c}_t \right), \mathbf{p}^i_t \in \mathcal{A} ,
\end{aligned}
\end{equation}
where $\mathbb{D}$ represents the function that calculates the distance between a point and the closest edge of the rectangle when the point lies inside the rectangle. 
It is defined as 
\begin{equation}
\begin{aligned}
\mathbb{D}\left( \mathbf{p}, \mathbf{c} \right) = 
\begin{cases}
\min\limits_{i\in\left\{bl,br,tl,tr\right\}} d_i, & \text{if } \mathbf{p} \in \mathbf{C}, \\
0, & \text{if } \mathbf{p} \notin \mathbf{C}.
\end{cases}
\end{aligned}
\end{equation}
where $d_i$ denotes the distance between the point and the four edges (bottom left, bottom right, top left, top right) of the rectangle $\mathbf{c}$, and $\mathbf{C}$ represents the area enclosed by the rectangle.

Additionally, to prevent the corridor from shrinking excessively, we introduce an area loss to encourage larger corridor sizes.
The area loss is defined as
\begin{equation}
\label{eq: area}
\begin{aligned}
    \mathcal{L}_{area} = \sum_{t=1}^{N} e^{-\alpha w_t l_t},
\end{aligned}
\end{equation}
where $\alpha$ is a scaling parameter that controls the magnitude of the penalty, and $w_t$ and $l_t$ represent the width and length of the rectangle at time $t$, respectively. 

\subsection{Corridor Refinement}
\label{sec: corref}

Leveraging the prediction and mapping results, we further refine the initially predicted corridor to mitigate potential conflicts with perception results.
The refinement process is also formulated as an MER problem, as described in Section \ref{sec: ann}, but with different data settings.
Specifically, the predicted agent boxes and the curbs are treated as obstacle points that the refined rectangle must exclude.
The refined rectangle retains the origin of the predicted rectangle $\left[ \hat{c}_x, \hat{c}_y \right]_t$, while its boundary is constrained by the predicted rectangle size $\left[ \hat{c}_l, \hat{c}_w \right]_t$. 
This ensures that the refinement does not introduce large positional offsets or excessive shape distortion.
By solving the MER problem at each timestep, the refined corridor $\mathcal{C}^{+}$ is obtained through this post-processing step, effectively “shrinks” the corridor to align with perceived obstacles.


\section{Differentiable Optimization with Corridor}
\label{sec:opt}
Tracking the reference trajectory using a vehicle model has been extensively studied in control theory, often formulated as an optimization problem. 
With the compact representation of our corridor, the safety constraint can be seamlessly integrated, framing the problem as a quadratic programming (QP).
Recent advancements\cite{amos2017optnet}\cite{agrawal2019differentiable} have made such optimization processes differentiable, allowing gradients to be backpropagated to the cost functions and constraints. 
This breakthrough inspires us to embed the optimization process into the network as a differentiable head. 
To the best of our knowledge, this is the first approach to integrate trajectory optimization into an end-to-end driving framework, thereby expanding the learnable components.
The forward and backward processes are described in the following sections.

\begin{table*}[t]
\centering
\resizebox{\linewidth}{!}{
\begin{tabular}{c|c|cccc|cccc|cccc|cccc}
\toprule
\multirow{2}{*}{ID} & \multirow{2}{*}{Method} & \multicolumn{4}{c|}{L2 (m) $\downarrow$} & \multicolumn{4}{c|}{ACR (\%) $\downarrow$} & \multicolumn{4}{c|}{CCR (\%) $\downarrow$ }& \multicolumn{4}{c}{Closed-loop Metric $\uparrow$} \\ 
 & & 1s & 2s & 3s & Avg. & 1s & 2s & 3s & Avg. & 1s & 2s & 3s & Avg. & DS* & SR* (\%)& DS$^\star$ & SR$^\star$ (\%) \\

\midrule
0 & ST-P3  & 1.59 & 2.64 & 3.73 & 2.65 & 0.69 & 3.62 & 8.39 & 4.23 & 2.53 & 8.17 & 14.4 & 8.37 &-&-&-&- \\
1 & UniAD  & 0.20 & 0.42 & 0.75 & 0.46 & 0.02 & 0.25 & 0.84 & 0.37 & 0.20 & 1.33 & 3.24 & 1.59 &37.72&9.54&47.01& 20.00 \\
2 & VAD-Base  & 0.17 & 0.34 & 0.60 & 0.37 & 0.04 & 0.27 & 0.67 & 0.33 & 0.21 & 2.13 & 5.06 & 2.47 &\textbf{39.42}&10.00&50.92&30.00 \\
3 & AD-MLP  & \textbf{0.15} & \textbf{0.32} & 0.59 & \textbf{0.35} & \textbf{0.00} & 0.27 & 0.85 & 0.37 & 0.27 & 2.52 & 6.60 & 2.93 &9.14&0.00&18.54& 0.00 \\
4 & BEV-Planner & 0.16 & \textbf{0.32} & \textbf{0.57} & \textbf{0.35} & \textbf{0.00} & 0.29 & 0.73 & 0.34 & 0.35 & 2.62 & 6.51 & 3.16 &-&-&-&- \\
\midrule
5 & CorDriver & 0.18 & 0.34 & 0.59 & 0.37 & 0.02 & 0.06 & 0.31 & 0.13 & 0.16 & 0.61 & 2.01 & 0.92 &37.53&\textbf{12.72}&\textbf{67.85}&\textbf{37.50} \\
6 & CorDriver$^{+}$ & 0.18 & 0.35 & 0.60 & 0.38 & \textbf{0.00} & \textbf{0.04} & \textbf{0.29} & \textbf{0.11} & \textbf{0.14} & \textbf{0.57} & \textbf{1.86} & \textbf{0.85} &-&-&-&- \\

\bottomrule

\end{tabular}
}
\captionsetup{font={small}}
\caption{Planning results on nuScenes. CorDriver utilizes the raw predicted corridor as optimization constraints, while CorDriver$^+$ further refines the corridor using perception results. Both models demonstrate exceptional performance in reducing collision rates.  In the closed-loop metrics, DS and SR represent Driving Score and Success Rate, respectively. The superscript * indicates tests on bench2drive220, while $^\star$ denotes tests on dev10.}
\label{tab:nusc_results}
\vspace{-0.7cm}
\end{table*}

\subsection{Forward Optimization}
\label{sec: foward}
The forward pass involves solving an optimization problem to compute the optimal control sequence that minimizes a defined cost function while respecting system dynamics and constraints. 
The optimization is formulated as follows:
\begin{equation}
\label{eq: qp}
\begin{aligned}
\min_{\mathbf{u}_0, \mathbf{u}_1, \dots, \mathbf{u}_{N-1}} & 
\sum_{t=0}^{N-1} 
\left[
    \hat{\mathbf{x}}_{t+1}^\top \mathbf{Q} \hat{\mathbf{x}}_{t+1} 
    + \mathbf{u}_t^\top \mathbf{R} \mathbf{u}_t
\right] \\
\text{s.t.} \quad 
\mathbf{x}_{t+1} & = \mathbf{A} \mathbf{x}_t + \mathbf{B} \mathbf{u}_t, \\
\mathbf{x}_t & \in \mathbf{C}_t, \quad \forall t \in [1, N], \\
\mathbf{u}_t & \in [\mathbf{u}_{min}, \mathbf{u}_{max}], \quad \forall t \in [0, N-1], \\
\mathbf{x}_0 & = \mathbf{x}_{init}.
\end{aligned}
\end{equation}

Here, $\mathbf{x}=\left[p_x, p_y, \theta, v\right]^\top \in \mathbb{R}^4$ includes the position, heading, and speed of the vehicle, and the control vector $\mathbf{u}=\left[ a, \delta \right] \in \mathbb{R}^2$ represents the acceleration and steering angle.
The bounds $\mathbf{u}_{min}$ and $\mathbf{u}_{max}$ define the feasible range for the control inputs, 
and $\mathbf{x}_{init}$ specifies the initial state of the vehicle.

The cost function comprises two component, the tracking cost $\hat{\mathbf{x}}_{t+1}^\top \mathbf{Q} \hat{\mathbf{x}}_{t+1}$ and the control effort $\mathbf{u}_t^\top \mathbf{R} \mathbf{u}_t$, weighted by the positive diagonal matrices $\mathbf{Q}$ and $\mathbf{R}$.
Specifically, $\hat{\mathbf{x}}_t = \mathbf{x}_{t} - \hat{\boldsymbol{\xi}}_{t}$ denotes the deviation from the reference trajectory $\hat{\boldsymbol{\xi}}_{t}$.

The dynamics of the vehicle are modeled using a linearized bicycle kinematic model. The discrete-time dynamics are given by
$\mathbf{x}_{t+1} = \mathbf{A} \mathbf{x}_t + \mathbf{B} \mathbf{u}_t$
where the matrices $\mathbf{A}$ and $\mathbf{B}$ are defined as 
\begin{equation}
\begin{aligned}
\mathbf{A} = & 
\begin{bmatrix}
1 & 0 & -v \sin\theta \Delta t & \cos\theta \Delta t \\
0 & 1 & v \cos\theta \Delta t & \sin\theta \Delta t \\
0 & 0 & 1 & \frac{\tan\theta}{L} \Delta t \\
0 & 0 & 0 & 1
\end{bmatrix} , \\
\mathbf{B} = & 
\begin{bmatrix}
0 & 0 \\ 
0 & 0 \\ 
0 & \frac{v}{L\cos^2\theta} \Delta t\\ 
\Delta t & 0
\end{bmatrix}, \\
\end{aligned}
\end{equation}
where $\Delta t$ is the time step, $L$ the wheelbase of the vehicle.

The constraint $\mathbf{x}_t \in \mathbf{C}_t$ ensures that the ego vehicle remains within the designated corridor.
Specifically, the predicted corridor $\hat{\mathcal{C}}$ can be converted into its $\mathcal{H}$-representation, 
which characterizes convex polytopes (rectangles in our case) as intersections of half-spaces, each defined by a linear inequality.
Consequently, constraining a point $\mathbf{p}$ to lie inside a rectangle is expressed as
\begin{equation}
    \mathbf{A}_{\mathbf{c}}\mathbf{p} \leq \mathbf{b}_{\mathbf{c}} ,
\end{equation}
where $\mathbf{A}_{\mathbf{c}}$ and $\mathbf{b}_{\mathbf{c}}$ are the inequality coefficient converted from the rectangle $\mathbf{c}_t$ at each timestamp.

Additionally, the size of the ego vehicle should be considered in the constraint. 
By projecting the vertices of the ego vehicle into the world frame, the corridor constraint is further detailed as
\begin{equation}
\begin{aligned}
    \mathbf{A}_{\mathbf{c}}
    \left(\bar{\mathbf{R}}_{\theta}
\begin{bmatrix}
l_{ego} & -w_{ego} \\ 
l_{ego} & w_{ego} \\ 
-l_{ego} & w_{ego}\\ 
-l_{ego} & -w_{ego}
\end{bmatrix}^\top
+ \mathbf{P}_{xy}
    \right)
    \leq \mathbf{b}_{\mathbf{c}} ,
\end{aligned} 
\end{equation} 
where $\bar{\mathbf{R}}_{\theta}$ is the linearized rotation matrix from the current ego heading $\theta$, and $\mathbf{P}_{xy}$ is the stacked position matrix of the ego vehicle.
The parameters $l_{ego}$ and $w_{ego}$ represent the half length and width of the ego vehicle, respectively.

Finally by solving the QP, we obtain the optimal control sequence $\mathbf{u}$. Forward propagating $\mathbf{u}$ from the initial state $\mathbf{x}_0$ leads to the final trajectory $\widetilde{\boldsymbol{\xi}}$.
The imitation loss computes the deviation from the human driving trajectory $\boldsymbol{\xi}^{\star}$.
\begin{equation}
\label{eq: ctrl}
\begin{aligned}
    \mathcal{L}_{imi} & = \mathcal{L}_1\left( \widetilde{\boldsymbol{\xi}} , \boldsymbol{\xi}^{\star} \right) . 
\end{aligned}
\end{equation}

By combining all the weighted losses, namely, the corridor loss \eqref{eq: cor}, map safety loss \eqref{eq: map}, agent safety loss \eqref{eq: agent}, area loss \eqref{eq: area} and imitation loss \eqref{eq: ctrl} together with the losses from VAD, we formulate the overall learning loss.
Notably, the planning constraints in VAD are no longer required, as our approach introduces a more concrete and compact representation for the trajectory constraint.

\subsection{Backward Gradient Propagation}
The backward pass propagates gradients from the QP solution to the problem parameters by leveraging implicit differentiation of the Karush-Kuhn-Tucker (KKT) conditions\cite{amos2017optnet}.
The KKT conditions relate the optimal solution to the problem's parameters, enabling gradient flow through the optimization layer. 
By solving the linearized KKT system, gradients are efficiently computed, allowing the optimization layer to adjust control policies, planning trajectories and corridor constraints. 
 A detailed discussion of these settings is provided in Section \ref{sec: abla QP}.

\begin{table*}[ht]
\centering
\begin{tabular}{cl|cccc|cccc|cccc}
\toprule

\multirow{2}{*}{Model} & \multirow{2}{*}{Description}  & \multicolumn{4}{c|}{L2 (m) $\downarrow$} & \multicolumn{4}{c|}{ACR (\%) $\downarrow$} & \multicolumn{4}{c}{CCR (\%) $\downarrow$}  \\
 & & 1s & 2s & 3s & Avg. & 1s & 2s & 3s & Avg. & 1s & 2s & 3s & Avg.  \\
\midrule
 $\mathcal{M}_0$ & Base & 0.17 & 0.37 & 0.70 & 0.41 & 0.0 & 0.18 & 1.00 & 0.39 & 0.17 & 1.06 & 2.87 & 1.37 \\
 $\mathcal{M}_1$ & $\mathcal{M}_0$ + $\mathcal{L}_{cor}$ & 0.17 & 0.37 & 0.69 & 0.41 ($\downarrow$0.0\%) & 0.04 & 0.25 & 0.78 & 0.36 ($\downarrow$7.7\%) & 0.10 & 1.00 & 2.87 & 1.32 ($\downarrow$3.6\%)\\
 $\mathcal{M}_2$ & $\mathcal{M}_1$ + Aux Loss  & 0.16 & 0.33 & 0.59 & 0.36 ($\downarrow$12.2\%) & 0.0 & 0.14 & 0.66 & 0.27 ($\downarrow$30.8\%) & 0.12 & 0.88 & 2.85 & 1.28 ($\downarrow$6.6\%) \\
 $\mathcal{M}_3$ & $\mathcal{M}_2$ + Opt & 0.18 & 0.36 & 0.62 & 0.39 ($\downarrow$4.9\%) & 0.0 & 0.12 & 0.55 & 0.22 ($\downarrow$43.6\%) & 0.14 & 0.78 & 2.38 & 1.10 ($\downarrow$19.7\%) \\
\bottomrule
\end{tabular}
\captionsetup{font={small}}
\caption{Ablation study on each component. 
The base model $\mathcal{M}_0$ only trains the planning trajectory. 
$\mathcal{M}_1$ introduces corridor learning using only $\mathcal{L}_{cor}$. $\mathcal{M}_2$ extends this by incorporating auxiliary losses $\mathcal{L}_{map}$, $\mathcal{L}_{agent}$ and $\mathcal{L}_{area}$. 
$\mathcal{M}_3$  utilizes the reference trajectory and corridor from $\mathcal{M}_2$ to generate the optimized final trajectory. }
\label{tab:abla}
\vspace{-0.5cm}
\end{table*}


\section{Experiments}
\label{sec:validation}

\subsection{Implementation Details}

Our training process is divided into two stages.
In the first stage, the differentiable optimization is excluded, and the remaining tasks — detection, prediction, mapping, planning, and corridor prediction — are trained for 48 epochs.
This step ensures a reasonably accurate trajectory and corridor prediction, which is necessary to enable the optimization process in the next stage.
In the second stage, spanning 12 epochs, the imitation loss is incorporated while the perception heads are frozen.
If the optimization fails during this stage, the reference trajectory $\hat{\boldsymbol{\xi}}$ is used to compute the imitation loss $\mathcal{L}_{imi}$.
The model is trained on 4 A100 GPUs with a batch size of 2 per GPU, requiring approximately 4 days for the first stage and less than 1 day for the second stage.

The hyperparameters are set as follows.
Following the convention in most prior works, we set $N=6$ with a gap of $\Delta t=0.5s$ between each timestamp, resulting in a planning horizon of 3 seconds.
The BEV range is configured to 60m $\times$ 30m, with a resolution of 0.15m.
For corridor generation, we set ego trajectory during $T_{ego}=\left[-5s, +5s\right]$ when selecting lanes as obstacles.
The maximum length $l_{max}$ and width $w_{max}$ of the boundary are set to 30m and 15m, respectively, while the obstacle threshold is set to $\delta_{obs}=0.5m$.
For the area loss in equation \eqref{eq: area}, $\alpha$ is set to 0.01.
The weights for all introduced losses are set to 1.0.

During inference, in cases where the optimization fails, we employ a soft-constrained variant that relaxes the corridor inequality constraints using slack variables, while all other settings remain as described in Section \ref{sec:opt}. 
This fallback mechanism has proven effective, as no failures have been observed in our test cases, ensuring the robustness of our method.

\subsection{Main Results}

Previous works \cite{Li_2024_CVPR}\cite{weng2024drive} have revealed limitations in existing evaluations, such as coarse spatial resolution and lack of ego orientation considerations.
To overcome these shortcomings, we adopt the evaluation pipeline from BEV-Planner\footnote{\url{https://github.com/NVlabs/BEV-Planner}} to assess planning performance.
To be detailed, the ego vehicle, agents, and curbs are projected onto a BEV image with a finer grid size of 0.1m.
The Agent Collision Rate (ACR) and Curb Collision Rate (CCR) are calculated by counting intersected pixels.
A trajectory is considered in collision if an intersection occurs at any timestamp, and the L2 metric is averaged over time, consistent with VAD \cite{jiang2023vad}.


We evaluate our algorithm on the public nuScenes dataset \cite{caesar2020nuscenes}, a large-scale multimodal benchmark widely used to advance autonomous driving research.
The results of our evaluation are presented in Table \ref{tab:nusc_results}.
For a fair comparison, UniAD, VAD-Base, and AD-MLP are incorporated with the ego status in the planning stage.
To evaluate our method in closed-loop settings, we integrate it into Bench2Drive \cite{jia2024bench2drive}, a comprehensive benchmark for assessing closed-loop end-to-end driving. Bench2Drive provides a standardized framework for fair comparisons within CARLA v2, encompassing diverse driving scenarios under various weather conditions and locations. Evaluations are conducted in two settings: the bench2drive220 split under protocol v0.0.1 and the dev10 split under protocol v0.03. The driving score and success rate are recorded.

In the open-loop setting on nuScenes, our proposed method demonstrates outstanding performance in reducing collision rates. Specifically, the base model, CorDriver, achieves an average reduction in object collisions by 60.6\% compared to the leading model, VAD, and a 42.1\% reduction in curb collisions compared to UniAD.
With the additional refinement introduced in Section \ref{sec: corref}, the enhanced version, CorDriver$^{+}$, further achieves an impressive averaged 0.11\% ACR and 0.85\% CCR, corresponding to a 66.7\% and 46.5\% decrease in agent and curb collisions, respectively.
In the closed-loop experiments, our method achieves higher success rates, demonstrating its ability to keep the vehicle within lanes and successfully complete the route. With driving scores competitive with existing methods, CorDriver maintains safe and efficient driving behavior without increasing infractions. These results highlight the advantages of corridor learning and planning in reducing collisions and enhancing driving robustness.

\begin{figure}[ht]
	\centering
	\includegraphics[width=0.95\linewidth]{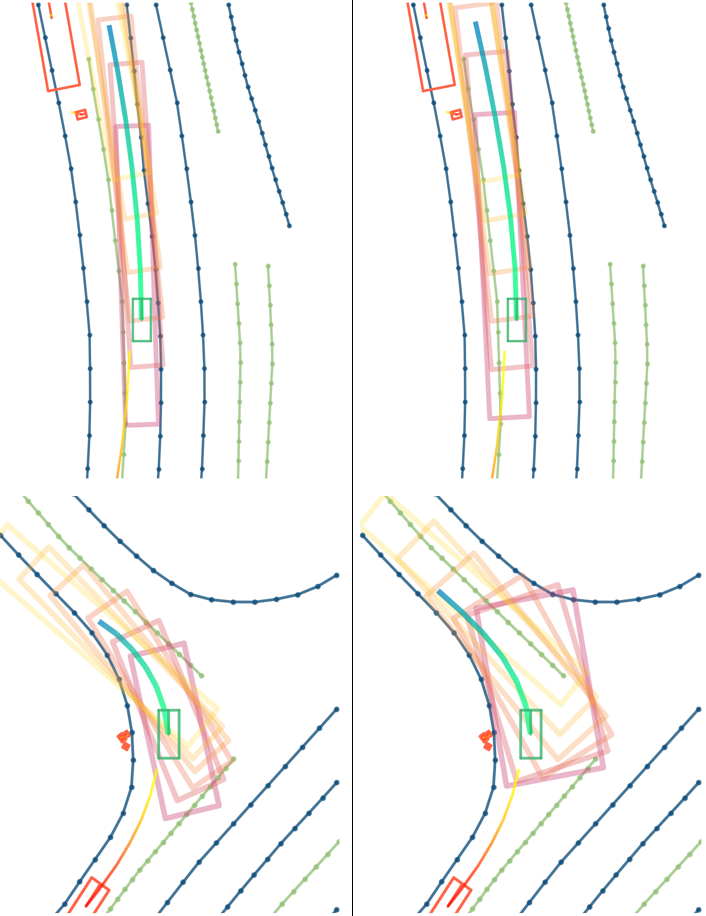}
	\captionsetup{font={small}}
	\caption{Visualization comparison of the learned corridors with auxiliary losses. Left: Corridors learned using only $\mathcal{L}_{cor}$. Right: Auxiliary losses are incorporated, which reduces overlap with curbs.  Ground-truth maps and agents are displayed for clearer comparison. Minor intersection between the predicted corridor and ground-truth map may occur due to imperfections in perception outputs, also seen in Fig \ref{fig:qualitative}.  Note that the bottom case follow the left-hand traffic rules. }
	\vspace{-0.8cm}
	\label{fig:cor loss}
\end{figure}

\begin{figure*}[h]
	\centering
	\includegraphics[width=1.0\linewidth]{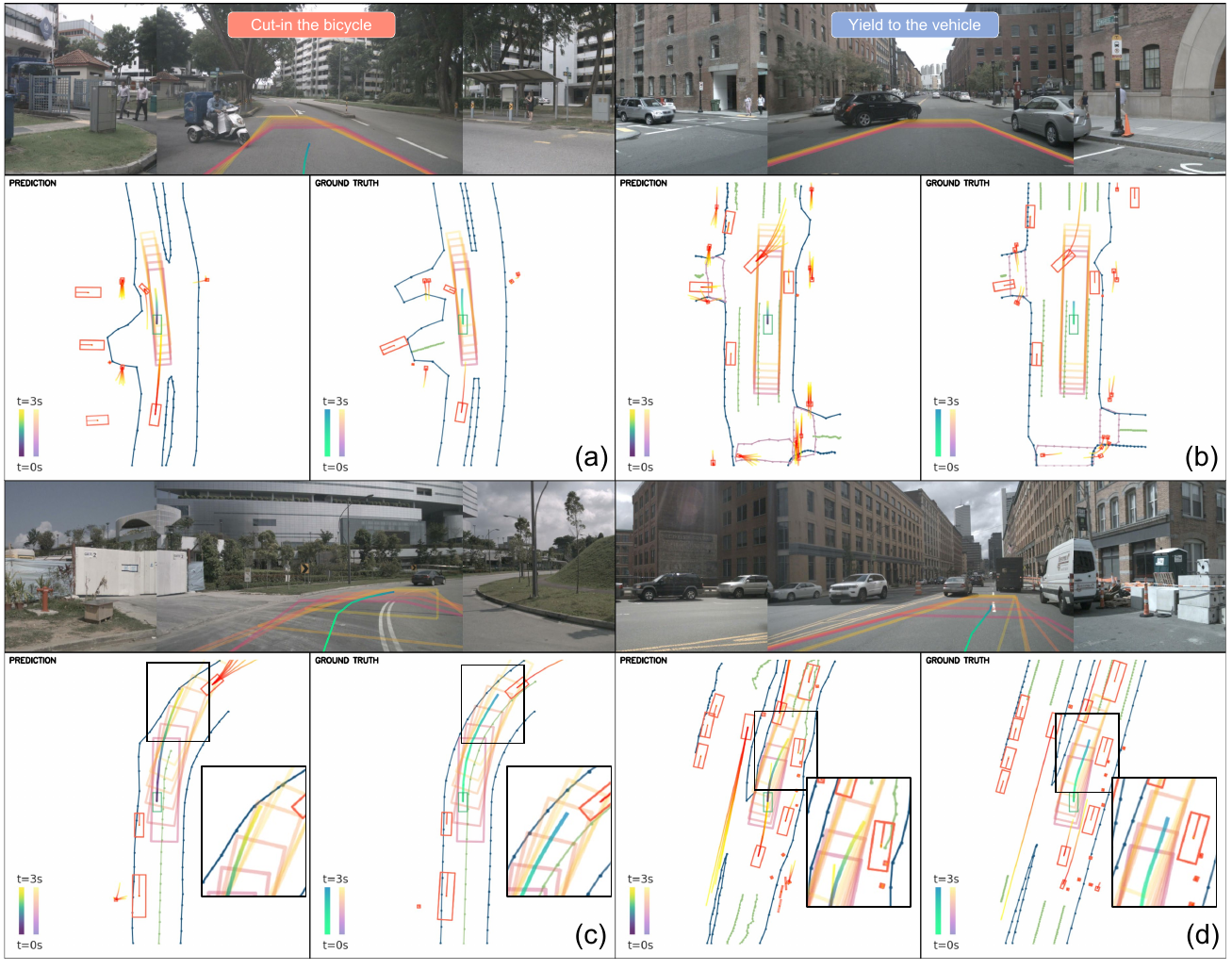}
	\captionsetup{font={small}}
	\caption{Qualitative Results. Each subfigure presents the perspective view with projected trajectory and corridor, and the 'PREDICTION' results (perception outputs with the \textbf{reference} trajectory) alongside the 'GROUND TRUTH' (ground-truth maps and agents and the \textbf{optimized} trajectory).
    Subfigures (c) and (d)
    \color{black} highlight cases where the reference trajectory collides, but the corridor constraints successfully guide the optimized trajectory to remain safe. Slight discrepancies between the perspective and BEV view may occur due to the estimated corridor height.}
	\vspace{-0.6cm}
	\label{fig:qualitative}
\end{figure*}

\subsection{Ablation on Modular Designs}
\label{sec: ablation}
We compare the effects of different modules, as summarized in Table \ref{tab:abla}.
For efficient evaluation, the models are trained for 12 epochs, building on a pre-trained perception model. 
The validation results show that simply incorporating the corridor learning task yields comparable trajectory precision with a modest reduction in collision rates.
However, the introduction of auxiliary losses significantly improves the accuracy of corridor predictions, which in turn enhances both the quality and safety of the predicted trajectory.
This implies a potential benefit of recognizing driving areas when planning the ego trajectory.
Differences in corridor predictions with and without auxiliary losses are visualized in Fig. \ref{fig:cor loss}.
Additionally, while optimization introduces some precision loss due to modeling limitations, it enforces safety constraints, leading to a substantial reduction in collisions with agents and curbs. We consider this trade-off worthwhile, as minimizing collisions is more critical in practice.
The optimization has a more pronounced effect on reducing curb collisions, demonstrating that the corridor effectively captures road geometries and constrains the reference trajectory from deviating off-road.
These findings highlight the importance of enforcing explicit constraints in trajectory optimization, showcasing the effectiveness of our interpretable planning process in improving overall safety.
\begin{table*}[ht]
\centering
\begin{tabular}{c|ccc|cccc|cccc|cccc}
\toprule
\multirow{2}{*}{ID} &  \multirow{2}{*}{weight}& \multirow{2}{*}{corridor} &\multirow{2}{*}{trajectory} & \multicolumn{4}{c|}{L2 (m) $\downarrow$} & \multicolumn{4}{c|}{ACR (\%) $\downarrow$} & \multicolumn{4}{c}{CCR (\%) $\downarrow$}  \\
 &  & && 1s & 2s & 3s & Avg. & 1s & 2s & 3s & Avg. & 1s & 2s & 3s & Avg.  \\

\midrule
0 & & &  & 0.178 & 0.342 & 0.595 & \cellcolor{orange!50} 0.372 &  0.0195 & 0.0781 & 0.391 & 0.163 & 0.156 & 0.684 & 2.012 & \cellcolor{orange!50} 0.951\\
\rowcolor[gray]{0.9} 1 & \Checkmark & & & 0.179 & 0.342 & 0.597 & \cellcolor{yellow!50} 0.373 & 0.0195 & 0.0586 & 0.313 & \cellcolor{orange!50} 0.130 & 0.156 & 0.606 & 2.012 & \cellcolor{red!50} 0.925  \\
2 & & \Checkmark & & 0.179 & 0.341 & 0.596 & \cellcolor{orange!50} 0.372 & 0 & 0.0391 & 0.410 & 0.150 & 0.195 & 0.703 & 2.168 &  1.022 \\
3 & & & \Checkmark & 0.182 & 0.348 & 0.607 & 0.379 & 0.0195 & 0.0977 & 0.352 & 0.156 & 0.176 & 0.957 & 2.442 & 1.192 \\
4 & \Checkmark& \Checkmark& & 0.179 & 0.343 & 0.602 & 0.375 & 0 & 0.0586 & 0.430 & 0.163 & 0.195 & 0.606 & 2.208 & \cellcolor{yellow!50}{1.003} \\
5 & \Checkmark& &\Checkmark& 0.181 & 0.346 & 0.600 & 0.376 & 0.0195 & 0.0977 & 0.332 & 0.150 & 0.195 & 0.821 & 2.305 & 1.107 \\
6 & & \Checkmark& \Checkmark& 0.175 & 0.344 & 0.610 & 0.376 & 0 & 0.0391 & 0.313 & \cellcolor{red!50} 0.117 & 0.156 & 0.684 & 2.872 & 1.237 \\
7 & \Checkmark& \Checkmark& \Checkmark& 0.175 & 0.337 & 0.594& \cellcolor{red!50} 0.369 & 0 & 0.0195& 0.391& \cellcolor{yellow!50} 0.137 & 0.156 & 0.664 & 2.266 & 1.029\\
\bottomrule
\end{tabular}
\captionsetup{font={small}}
\caption{Ablation results of learnable parameters in the optimization. The top 3 averaged metrics are marked with \colorbox{red!50}{red}, \colorbox{orange!50}{orange} and \colorbox{yellow!50}{yellow}.  Setting ID 1 achieves best overall performance.} 
\label{tab: abla QP}
\vspace{-0.7cm}
\end{table*}


\subsection{Ablation on Optimization Learning}
\label{sec: abla QP}

This section investigates the impact of gradient propagation in the optimization process.
Starting with the trained model after the first stage, which is capable of predicting both the corridor and the trajectory, we incorporate $\mathcal{L}_{imi}$ into training to refine these predictions.
Specifically, the weight matrices $\mathbf{Q}$ and $\mathbf{R}$ of the QP formulation are treated as learnable parameters.
The gradient of $\mathcal{L}_{imi}$ is selectively detached to isolate its effects on the weight matrices, the reference trajectory $\hat{\boldsymbol{\xi}}$ and corridor constraints $\hat{\mathcal{C}}$.

The results shown in Table \ref{tab: abla QP} indicate that training only the weight matrices yields the best overall performance, while introducing learnable trajectory and corridor components leads to comparable or even degraded performance.
Our analysis provides the following insights. During training, we observe larger gradient oscillations for the trajectory and corridor components compared to the weight matrices, which primarily contribute to poorer training outcomes. 
These fluctuations arise because the influence of trajectory and corridor varies significantly across different scenarios. For example, when the corridor is spacious, its impact on the optimized trajectory is minimal, whereas tighter constraints force larger control adjustments, leading to more pronounced changes in the optimization result.
In contrast, the weight matrices regulate cost term importance, providing more stable influence on the optimization across samples.
This smoother gradient behavior is helpful for convergence and performance.

\subsection{Other Results}
We present visualization results in several challenging scenarios, such as interacting intersections, high-curvature turns and merging into traffic, as shown in Fig. \ref{fig:qualitative}. More cases are available on the project page.
Besides, we measure the inference time of our model on an NVIDIA A100 GPU. The multi-task backbone requires an average of 159.6 ms per frame, nearly identical to the base VAD model. In contrast, the optimization module adds an extra 44.7 ms per frame, increasing the total computation time by 28\% to approximately 200 ms, well above the typical real-time threshold of 30 ms per frame. Further acceleration would be required for deployment on real-world vehicles, particularly in high-speed scenarios.

\section{Conclusion}
\label{sec:Conclusion}
This work enhances the safety of end-to-end autonomous driving by using corridors as a planning representation. Incorporating corridor prediction as a constraint in trajectory optimization increases both safety and interpretability.
We also explore the role of differentiable optimization within the end-to-end framework, demonstrating improvements through training certain components.
Additionally, our findings reveal that unstable gradients from the optimization process pose a challenge to effective learning. 
Addressing this issue might involve techniques such as smoothing gradients through penalty functions instead of using hard constraints.
How to effectively include the model-based optimizations into scalable end-to-end approaches remains a promising direction for future research.

\newlength{\bibitemsep}\setlength{\bibitemsep}{0.00\baselineskip}
\newlength{\bibparskip}\setlength{\bibparskip}{0pt}
\let\oldthebibliography\thebibliography
\renewcommand\thebibliography[1]{
	\oldthebibliography{#1}
	\setlength{\parskip}{\bibitemsep}
	\setlength{\itemsep}{\bibparskip}
}
\bibliography{references}

\end{document}